\documentclass[prl,reprint]{revtex4-1}

\usepackage{savesym}
\usepackage{amsmath}    
\usepackage{amssymb}    

\usepackage{graphicx}   
\usepackage{color}     
\usepackage{bm}  
\usepackage{bbold}
\usepackage{subfigure} 
\usepackage{hyperref} 
\hypersetup{bookmarks=true, unicode=false, pdftoolbar=true, pdfmenubar=true, pdffitwindow=false, pdfstartview={FitH}, pdfnewwindow=true, colorlinks=true, linkcolor=black, citecolor=blue, filecolor=magenta, urlcolor=blue}

\usepackage{cases}

\newcommand{\e}{\mathrm{e}}
\newcommand{\dx}{\mathrm{d}x}
\newcommand{\dy}{\mathrm{d}y}
\newcommand{\sBethe}{s_\textrm{Bethe}}
\newcommand{\SBethe}{S_\textrm{Bethe}}

\newcommand{\UBethe}{U}
\newcommand{\sannealed}{s_\textrm{ann}}

\begin{document}
\title{Belief propagation for permutations, rankings, and partial orders}
\author{George T. Cantwell}
\affiliation{Santa Fe Institute, 1399 Hyde Park Road, Santa Fe, New Mexico 87501, USA}
\author{Cristopher Moore}
\affiliation{Santa Fe Institute, 1399 Hyde Park Road, Santa Fe, New Mexico 87501, USA}

\begin{abstract}
Many datasets give partial information about an ordering or ranking by indicating which team won a game, which item a user prefers, or who infected whom. We define a continuous spin system whose Gibbs distribution is the posterior distribution on permutations, given a probabilistic model of these interactions. Using the cavity method we derive a belief propagation algorithm that computes the marginal distribution of each node's position. In addition, the Bethe free energy lets us approximate the number of linear extensions of a partial order and perform model selection between competing probabilistic models, such as the Bradley-Terry-Luce model of noisy comparisons and its cousins.
\end{abstract}

\maketitle

Ranking or ordering objects is a natural problem in many contexts.
Mathematically, this task corresponds to finding ``good'' permutations of a finite set, or more generally, sampling from a distribution of good permutations. This can be surprisingly difficult. 

For example, suppose we observe a set of pairwise interactions, such as competitions, preferences, or conflicts, each of which is evidence that one object is ranked above another, and our goal is to rank them from strongest to weakest. Similarly, we might want to reconstruct the order in which nodes joined a growing network~\cite{young_phase_2019, navlakha_archaeo_2011}, for instance in an epidemic where contact tracing suggests links where one individual infected another. In cases like these, finding a permutation which minimizes the number of violations where the ordering goes the ``wrong'' way is NP-hard, i.e., among the hardest optimization problems in computer science~\cite{miller_reducibility_1972}. Even when there exist permutations consistent with all observed interactions, counting the number of such permutations, or computing the average positionof a given object, is \#P-complete~\cite{brightwell_counting_1991,brightwell_counting_1991-1}. Thus all these problems are believed to take exponential time in the worst case.

Pairwise comparisons can be represented as a directed graph $G$ whose edges $(i,j)$ indicate that $i \prec j$, i.e., $i$ ``beat'' $j$ and is therefore probably ranked above $j$. We assume a generative model: given a ground-truth permutation $\bm{\pi}$, we observe $G$ with probability $P(G \vert \bm{\pi})$ \footnote{Our model does not attempt to explain which pairs interact, only the outcome of these interactions.}.
If all permutations are equally likely a priori, and if we observe each $i \prec j$ independently with probability $f(\pi_i,\pi_j)$, the posterior has the form
\begin{equation}
	\label{eq:posterior}
	P(\bm{\pi} \vert G) = \frac{\prod_{(i,j) \in G} f( \pi_i, \pi_j )}{\sum_{\bm{\pi}'}\prod_{(i,j) \in G} f( \pi_i', \pi_j' )}.
\end{equation}
The framework of Eq.~\eqref{eq:posterior} may seem restrictive but we will see that it covers several interesting problems. Specifically, we consider (1)~counting linear extensions of partial orders; (2)~inferring the order in which a network grew; (3)~finding minimum feedback arc sets;  (4)~parameter estimation and model selection for rankings.

To advance an analogy with statistical physics, we interpret Eq.~\eqref{eq:posterior} as a Gibbs distribution $P(\bm{\pi}\vert G)=e^{-\beta H(\bm{\pi})}/Z$ at temperature $\beta^{-1}$ with Hamiltonian $H(\bm{\pi}) = \sum_{(i,j) \in G} h(\pi_i, \pi_j )$.
Of particular interest is the step function Hamiltonian
\begin{equation}
	h(\pi_i, \pi_j) = \Theta(\pi_i-\pi_j) =
	\begin{cases} 
		1 & \pi_i \geq \pi_j \\
		0 & \pi_i < \pi_j \, . 
	\end{cases}
	\label{eq:step}
\end{equation}
In this case, the energy $H(\bm{\pi})$ is the number of violations, i.e., the number of edges in $G$ that are oriented contrary to the ordering of the nodes in permutation $\bm{\pi}$ \cite{thompson_rankings_1964,park_diagrammatic_2010}.

We begin our investigation of the Hamiltonian~\eqref{eq:step} at zero temperature.
If the system is not frustrated, i.e. if there is a $\bm{\pi}$ for which $H(\bm{\pi})=0$, then we can view each directed edge $(i,j) \in G$ as a hard constraint that demands $\pi_i < \pi_j$.
This implies that $G$ is acyclic and defines a partial order: that is, a structure where we are given that $i \prec j$ for some pairs $i,j$. Partial orders are transitive: $i \prec j$ and $j \prec k$ implies $i \prec k$.
However, there may be pairs $i,j$ where neither $i \prec j$ nor $j \prec i$ is necessarily true, leaving their relative order ambiguous. In contrast, a \emph{total} or \emph{linear} order is a permutation---every object has an unambiguous rank, and all pairs of items are comparable. A \emph{linear extension} of a partial order is a total order that satisfies all the constraints of the partial order; it is is equivalent to a topological ordering of the corresponding directed acyclic graph. 
There are typically many such orders. 

Counting the linear extensions of a partial order is a well-known problem in computer science. Counting them exactly is \#P-complete~\cite{brightwell_counting_1991-1}, making it as hard as computing spin glass partition functions or matrix permanents, and almost certainly requiring exponential time. There are polynomial-time Monte Carlo algorithms~\cite{dyer_random_1991,karzanov_conductance_1991,lovasz_random_1993} but these are fairly slow in practice~\cite{talvitie_2018}. Here we provide a fast approximate algorithm based on belief propagation and the cavity method in a related spin system.

Permutation-valued states cause several challenges for the cavity method. First, \emph{a priori} the entropy grows super-extensively as $\log n! \sim n \log n$, creating a rather odd thermodynamic limit. Secondly, since each site can be in one of $n$ different states, we have something like a $q$-state Potts model where $q=n$, making the fields $n$-dimensional. Thirdly, no two sites can have the same state. This creates a global coupling, violating the local treelikeness that the cavity method assumes.

We can address all these problems with a rescaling to continuous variables. Rather than treat each $\pi_i$ as an integer from $1$ to $n$, we associate it with a real number $x_i$ in the unit interval. A state is a point in the $n$-dimensional unit hypercube, \mbox{$\bm{x} \in \left[ 0,1 \right]^n$}, which corresponds to the permutation given by the sorted order of its components. Since the $x_i$ are distinct with probability $1$, this removes the global coupling between sites.

\begin{figure}
    \centering
    \includegraphics[width=0.42\columnwidth]{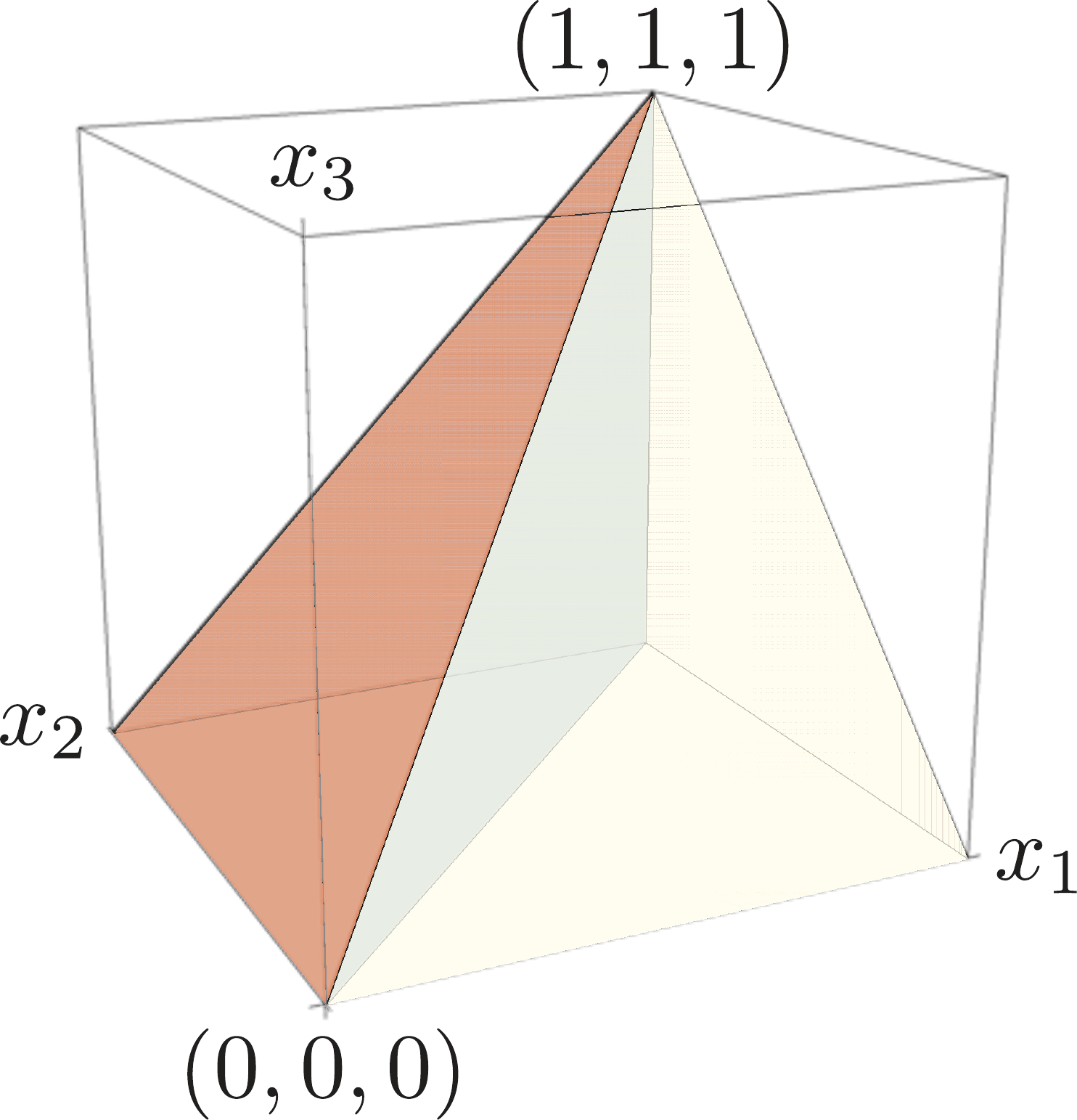}
    \caption{The convex polytope corresponding to the partial order $3 \prec 1, 3 \prec 2$, i.e., the subset of the unit cube where $x_3 < x_1$ and $x_3 < x_2$. It contains two simplices corresponding to the linear extensions $3 \prec 1 \prec 2$ and $3 \prec 2 \prec 1$, and has total volume $2/3!=1/3$.}
    \label{fig:poly}
\end{figure}

Under this rescaling, the set of linear extensions of a given partial order becomes a convex polytope whose facets correspond to its constraints. Figure~\ref{fig:poly} shows the polytope corresponding to the partial order $3 \prec 1, 3 \prec 2$, which has facets $x_3 < x_1$ and $x_3 < x_2$. This partial order has two linear extensions, $3 \prec 1 \prec 2$ and $3 \prec 2 \prec 1$. 

Although our variables $x_i$ live in the unit interval as opposed to the circle or sphere, we think of this rescaled model as a continuous spin system in the spirit of the XY model or the classical Heisenberg model~\cite{stanley_1968}.
At zero temperature the partition function $Z$ is the volume of the polytope of linear extensions. Since each permutation corresponds to a simplex with volume $1/n!$, this gives
\begin{equation}
\label{eq:rescaling}
\begin{aligned}
	Z &= \lim_{\beta \to \infty}  \int_{[0,1]^n} \prod_{(i , j) \in G} e^{-\beta \Theta(x_i-x_j)} \,\mathrm{d}\bm{x} \\ 	
	&= \frac{\text{\# linear extensions}}{n!} \, ,
\end{aligned}
\end{equation}
and because every linear extension is equally likely, the entropy is simply $S=\ln Z$. As we discuss below, this rescaling from $\{1,\ldots,n\}$ to the unit interval allows us to define a sensible thermodynamic limit where $Z$ behaves as a simple exponential and $S$ is linear in $n$.

Since counting linear extensions is \#P-complete, so is computing $Z$ or $S$ exactly. We will use the cavity method to approximate them. We start by pretending that the graph of comparisons is a tree---that it has no cycles even when the directions of the edges are erased. The distribution of the spins can then be factorized as
\begin{equation}
	P(\bm{x} \vert G) 
	= \frac{\prod_{(i,j) \in G} \mu_{ij}(x_i,x_j)}
	{\prod_i \mu_i(x_i)^{d_i-1}} \, . 
	\label{eq:factorize}
\end{equation}
Here $\mu_{i}(x_i)$ is the marginal probability density for the spin at node~$i$, $\mu_{ij}(x_i,x_j)$ is the joint marginal for the spins at nodes~$i$ and $j$, and $d_i$ is the degree of node $i$, i.e., the number of objects to which it is compared.
The entropy $S = -\langle \ln P \rangle $ is then given by the \emph{Bethe entropy}
\begin{equation}
	\SBethe = \sum_{(i,j) \in G} S_{ij} -  \sum_i (d_i-1) S_i \, , 
	\label{eq:Bethe_entropy}
\end{equation}
where $S_i = -\big\langle \ln \mu_i \big\rangle $ and $S_{ij} = -\big\langle \ln \mu_{ij} \big\rangle$ are the entropies of the one- and two-point marginals respectively. 

\begin{figure*}
	\includegraphics[width=1.0\linewidth]{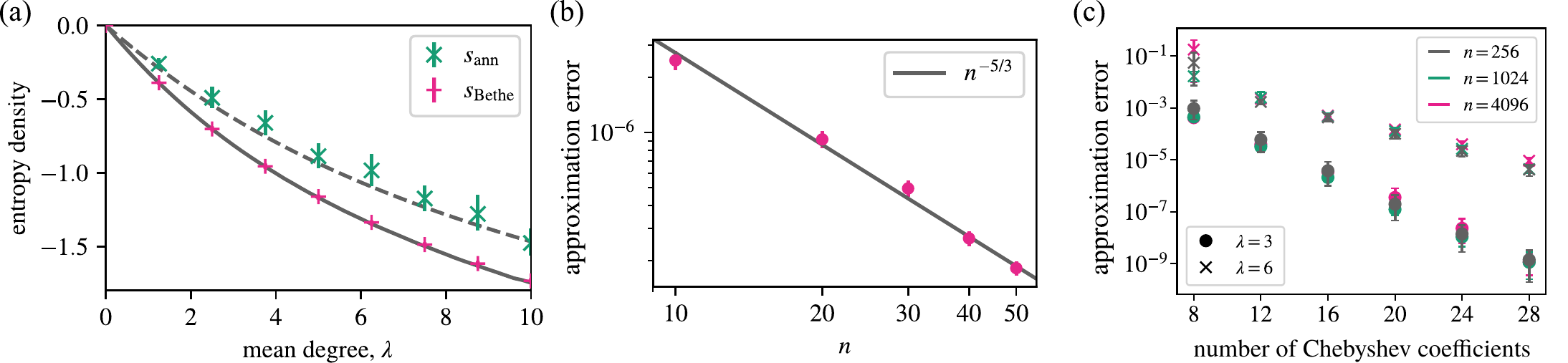}
	\caption{Entropy per site for random graphs. The number of linear extensions scales as $n! e^{sn}$ in the sparse case where $s < 0$, so after rescaling the spins to the unit interval as in Eq.~\eqref{eq:rescaling} the entropy is negative.
	(a) Values of $\sannealed$ and $\sBethe$ in random graphs of mean degree $\lambda$, estimated by belief propagation on random graphs of size $n=10^4$ under the assumption that $\ln Z$ is normally distributed. The dashed line is the analytic result Eq.~\eqref{eq:Gnp_annealed} for $\sannealed$, showing that our results are consistent with theory. The solid line is the value for $\sBethe$ given by population dynamics. 
	(b) Mean-squared error between $\sBethe$ and the exact value of $s = {1 \over n} \ln Z$ computed by exhaustive enumeration on random graphs of size $n \le 50$ and mean degree $\lambda=2$. Even when we approximate the messages with $d=32$ Chebyshev coefficients, $\sBethe$ rapidly converges to $s$, showing that the cavity method is asymptotically exact.
	(c) Mean-squared error for our estimate of $\sBethe$ as a function of the number $d$ of Chebyshev coefficients used in our approximation, for random graphs with $n$ nodes and mean degree $\lambda$. The error is calculated relative to $d=32$. There is clear dependence on $\lambda$ but not on $n$, and the estimate converges exponentially as $d$ increases.
		}
	\label{fig:Gnp_double}
\end{figure*}

These marginal distributions can be computed using belief propagation~\cite{pearl,mezard_information_2009}.
For each neighboring pair $(i,j)$, we ask what $j$'s marginal would be if $i$ were absent.
We denote this cavity marginal $\mu_{j \to i}(x)$, and think of it as a ``message'' or ``belief'' that $j$ sends to $i$. It in turn depends on the messages that $j$ receives from its neighbors $k$ other than $i$.
Note that if $i$ and $j$ are compared, messages go in both directions along the edge $(i,j)$.

For linear extensions, $\mu_{j \to i}(x_j)$ is proportional to the probability that $x_k < x_j$ (resp.\ $x_k > x_j$) for all $k \ne i$ such that $k \prec j$ (resp.\ $k \succ j$). Using the cumulative distribution functions
\begin{equation}
	M_{j \to i}(x_j) = \int_0^{x_j} 
	\mu_{j \to i}(y) \,\dy \, , 
\end{equation}
we can write this as
\begin{align}
	\mu_{j \to i}(x_j) 
	&\propto 
	\prod_{\substack{k \prec j \\ k \ne i}} 
	M_{k \to j}(x_j) 
	\times 
	\prod_{\substack{k \succ j \\ k \ne i}} 
	\big(1 - M_{k \to j}(x_j) \big) \, ,
	\label{eq:BP_message}
\end{align}
where we normalize so that $\int_0^1 \mu_{j \to i}(x_j) \,\dx_j = 1$. 
The one-point marginals are computed similarly, but using all of $j$'s neighbors, 
\begin{equation}
	\mu_j(x_j) 
	\propto 
	\prod_{k \prec j} 
	M_{k \to j}(x_j) 
	\times 
	\prod_{k \succ j} 
	\big(1 - M_{k \to j}(x_j) \big) \, ,
	\label{eq:BP_one_marginal}
\end{equation}
and the two-point marginal for an edge $i \prec j$ is
\begin{equation}
	\mu_{i j}(x_i, x_j) \propto \mu_{j \to i}(x_j) \,\mu_{i \to j}(x_i) \,\Theta(x_j - x_i) \, .
	\label{eq:BP_two_marginal}
\end{equation}
Note that we assume a uniform prior on the hypercube, and therefore a uniform prior on permutations.

This suggests an algorithm for counting linear extensions: solve the belief propagation equations~\eqref{eq:BP_message} by iterating until we reach a fixed point, and compute the entropy $S=\ln Z$ in Eq.~\eqref{eq:Bethe_entropy}.
If the graph of comparisons has loops, this algorithm is not exact, but the Bethe entropy is often an excellent asymptotic approximation to the true entropy. In particular, so long as the graph is sparse and locally treelike (with few short loops) we expect the resulting estimate of $Z$ to be correct up to subexponential terms.

Writing the system of equations in Eq.~\eqref{eq:BP_message} is one thing but solving it another, since it consists of a large system of nonlinear differential equations. Some further insights, however, reduce the complexity considerably. First, note that for any partial order the true marginal distribution $\mu_{i}(x)$ is a polynomial of degree at most $n-1$. 
To see this, recall that the unit hypercube can be divided into $n!$ simplices, each of which corresponds to one permutation. 
If $i$'s position in this permutation is $t$, then $t-1$ of the other spins $x_j$ must be less than $x_i$, and the other $n-t$ spins must be greater. The probability density for $x_i$ is thus proportional to $x^{t-1}(1-x)^{n-t}$, a so-called Bernstein polynomial.
Summing over all allowed permutations, $\mu_{i}(x)$ is a linear combination of such polynomials.
Similarly, the messages $\mu_{j \to i}(x)$ are polynomials of degree at most $n-2$.

This would allow us to solve Eq.~\eqref{eq:BP_message} by finding at most $n-1$ polynomial coefficients fonor the right-hand side.
However, this would require $O(n)$ computation for each edge.
Happily, this is unnecessary: we can approximate the messages as polynomials of lower degree using Chebyshev polynomials~\cite{trefethen2019approximation} $T_k(x)$, up to some maximum degree $d$, writing \mbox{$\mu_{j \to i}(x) \approx \sum_{k=0}^{d-1} c_{j \to i}^{\,k} T_k(x)$}. 
We initialize the messages to uniform distributions $T_0=1$, i.e., $\bm{c}_{j \to i} = \left( 1, 0, \dots, 0 \right)$.
We then iteratively update $\bm{c}_{j \to i}$ and $\bm{c}_{i \to j}$ for all edges $(i,j)$ using Eq.~\eqref{eq:BP_message}. 

Since the Chebyshev polynomial of degree $k$ on $[0,1]$ can be written $T_k(x) = \cos k\theta$ where $x = (1+\cos \theta)/2$, we can compute each update using the Fast Fourier Transform.
On sparse graphs, where the degree distribution has finite mean and variance, the computation time for an entire sweep is linear in the number of edges, and we typically converge to a fixed point in $O(\log n)$ sweeps.
In practice we obtain excellent results even when the polynomial degree $d$ is considerably smaller than $n$ (see Fig.~\ref{fig:Gnp_double}), allowing our method to scale easily to hundreds of thousands of nodes on a desktop computer.

As an initial test of our methods, suppose $G$ is a directed version of an Erd\H{o}s-R\'enyi graph $G_{n,p}$ where each pair of nodes is compared with probability $p$. To create a valid partial order, we label the nodes $1,\dots,n$ with a ground-truth permutation $\pi$, and orient the edges to agree with $\pi$. This model of a random partial order was studied in~\cite{alon_linear_1994}. For sparse graphs with average degree $\lambda$, i.e., in the limit $n \to \infty$ and $p=\lambda / n$, by applying results in Ref.~\cite{alon_linear_1994} we derive the annealed entropy per site  
\begin{equation}
	\sannealed 
	= \lim_{n \to \infty} \frac{\ln \langle Z_n \rangle}{n}
	= 1 - \ln \lambda \,+\! \int_0^1 \ln(1 - e^{-\lambda x}) \,\dx \, ,
	\label{eq:Gnp_annealed}
\end{equation}
so the expected number of linear extensions is $\langle n! Z_n  \rangle  = n! e^{n\sannealed}$ up to subexponential terms.

Note the unusual scaling of this problem. In sparse graphs where $\lambda=O(1)$ and there are $O(n)$ edges, each edge excludes a constant fraction of permutations, so the number of linear extensions is $n!$ multiplied by a simple exponential $\e^{sn}$ with $s < 0$. After rescaling to the unit interval as in~\eqref{eq:rescaling}, the total volume $Z$ is $\e^{sn}$, giving a valid thermodynamic limit where the entropy is extensive.
In contrast, in dense graphs with $O(n^2)$ edges, almost all of the $n!$ possible permutations are excluded, and the number of linear extensions is a simple exponential~\cite{alon_linear_1994}.

We conjecture that the cavity method is asymptotically exact in sparse random graphs: namely, that the typical quenched entropy is given by the Bethe entropy, $\lim_{n \to \infty} {1 \over n} \langle \ln Z_n \rangle = \lim_{n \to \infty}{1 \over n} \SBethe = \sBethe(\lambda)$. To test this conjecture, and measure the quality of our Chebyshev approximation, we use population dynamics~\cite{mezard_information_2009} to derive a fixed-point distribution of messages on sparse infinite graphs to estimate $\sBethe$, and also carried out belief propagation on finite graphs.

By running belief propagation on multiple realizations of finite random graphs, we estimated the mean $\mu$ and variance $\sigma^2$ of $\ln Z$. On the physical assumption that $\ln Z$ is normally distributed (which was proved for the dense case in~\cite{alon_linear_1994} and is easy to prove for the sparse case when $\lambda$ is sufficiently small) we have $\sBethe = \mu / n$ and $\sannealed = (\mu + {1\over2}\sigma^2)/n$. 

Figure~\ref{fig:Gnp_double}(a) shows that the resulting estimates of $\sannealed$ reproduce the analytic result Eq.~\eqref{eq:Gnp_annealed}, showing that our method is consistent with theory. We also find that $\sBethe$ from belief propagation on finite graphs agrees closely with the prediction from population dynamics.

We compared $\sBethe$ with the exact entropy $s = {1 \over n} \ln (\text{\# linear extensions}/n!)$, using a dynamic programming algorithm to enumerate permutations exhaustively, on random graphs of size up to $n=50$. Fig.~\ref{fig:Gnp_double}(b) shows that $\sBethe$ converges quickly to the true entropy as $n$ increases, as $n^{-\alpha}$ where $\alpha \approx 5/3$, suggesting that the cavity method is asymptotically correct.

The results of Fig.~\ref{fig:Gnp_double}(a,b) were obtained using Chebyshev polynomials of degree $d=32$. While the exact marginals have degree $n-1$, we observe exponential convergence in the number $d$ of coefficients used in our Chebyshev approximation. As Fig.~\ref{fig:Gnp_double}(c) shows, if we fix the average degree $\lambda$ of the graph, our estimate of $\sBethe$ converges rapidly as $d$ increases, and the error does not depend on $n$. Thus our approximation of the messages with low-degree polynomials is also highly accurate.

Although counting linear extensions of a partial order is an established problem computer science, it is less familiar in physics.
It is not so far removed, however, from inference in growing network models \cite{young_phase_2019,navlakha_archaeo_2011} such as preferential attachment \cite{de_solla_price_networks_1965, albert_statistical_2002}, which are among the most widely studied models in the physics of complex networks.
If $G_t$ is the network at time $t$, these models specify a transition probability $P(G_{t+1} \vert G_t)$ along with an initial condition $G_0$. Since we often only observe a snapshot of the network, or its final state, an interesting problem is to reconstruct the history of the network, i.e., the order in which its nodes were added. For trees, one can calculate the full distribution of possible ``histories'' by which an observed graph could have grown \cite{cantwell_inference_2021}.
Our methods here provide an approximate solution for general graphs.

\begin{figure}
	\includegraphics[width=0.9\linewidth]{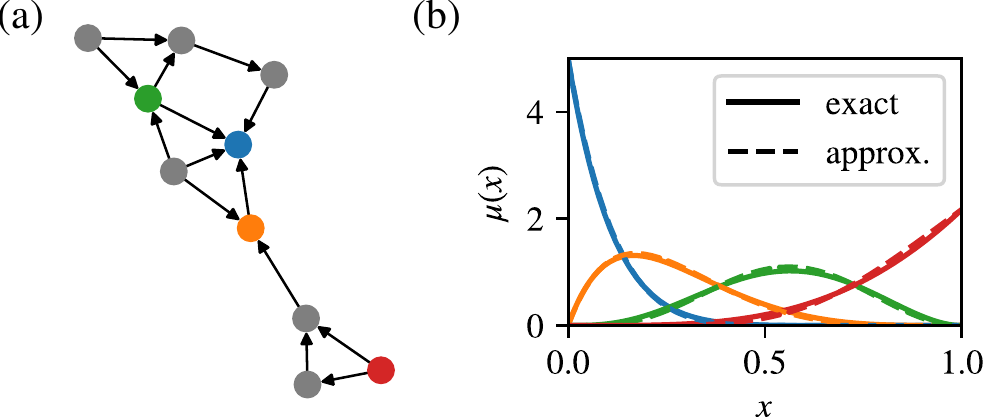}
	\caption{In (a) we show a randomly grown network with Poisson out-degree.  
	In (b) we show the posterior marginals for four representative nodes, colored to match (a), comparing those obtained by our method with the exact results of exhaustive enumeration. Despite the presence of short cycles, our belief propagation approach approximates the marginals quite closely, matching not just the means but the shapes of these distributions.
	}
	\label{fig:grown_graph}
\end{figure}

If the network's edges are directed, pointing from each new node to the node it attached to, they constitute a partial order with which the network's history must be consistent. We can then use Eq.~\eqref{eq:BP_one_marginal} to compute the posterior distribution of each node's arrival time. We show an example in Fig.~\ref{fig:grown_graph}. 
This graph is sufficiently small ($n=10$) that we can compare our method with an exhaustive enumeration of all $n!$ possible orderings. 
As shown in panel (b), the marginals we obtain are very accurate despite the presence of short loops. Thus we can efficiently approximate, not just the mean arrival time of each node, but its posterior distribution.

Next we turn to the frustrated case, where $G$ contains directed cycles such as $1 \prec 2 \prec 3 \prec 1$.
In this case, no permutation satisfies all the constraints and the data is inconsistent with $\beta = \infty$.
At finite $\beta$, Eq.~\eqref{eq:BP_message} becomes
\begin{align}
\mu_{j \to i}(x) 
& \propto \prod_{\substack{k \ne i}}
	 \int_{0}^{1}  \frac{\mu_{k \to j}(y) \,e^{-\beta g_{jk} \Theta(x-y)}}
	 {e^{-\beta \Theta(x-y)}+e^{-\beta \Theta(y-x)}}  \,\dy
	 \label{eq:BP_message_general} \\
& = \prod_{\substack{k \ne i}}
	 \int_{0}^{1} \frac{\mu_{k \to j}(y) \,e^{-\beta g_{jk} \Theta(x-y)}}
	 {1+e^{-\beta}} \,\dy
	 \label{eq:BP_message_stepfunction} 
\end{align}
where $g_{jk} = 1$ if $k \prec j$ and $-1$ if $k \succ j$. The marginals are again polynomials of degree at most $n-1$ since the density is a function only of $\bm{x}$'s permutation; we calculate these integrals as before using Chebyshev polynomials. 

The ground states of the Hamiltonian~\eqref{eq:step} are permutations that minimize the number of violated constraints.
These are known as minimum violation rankings~\cite{thompson_rankings_1964,park_diagrammatic_2010}.
Like counting linear extensions of a partial order, finding these rankings is computationally hard.
It is equivalent to the classic NP-complete ``minimum feedback arc set'' problem of making a directed graph acyclic by removing as few edges as possible~\cite{miller_reducibility_1972}.
Intuitively, this is because the landscape of rankings can be glassy, with multiple widely separated peaks. For instance, given the comparisons $1 \prec 2 \prec \cdots \prec n$, along with $n \prec 1$ and $1 \prec (n-1)$, then there are two minimum violation orderings, one where $n$ is first and another where it is last.

One strategy to find the frustrated ground states is to perform belief propagation at low temperature, where they dominate the Gibbs distribution. If we wish to find a single ground state regardless of its probability, we can use greedy decimation, removing the edge $(i,j)$ most likely to be violated according to $\mu_{ij}$ and iterating. Belief propagation for this problem was previously studied, using a discrete model, in Ref.~\cite{Zhao_2017}. 
We ran both our procedures on $40$ random directed graphs with $n=1000$ and mean degree $\lambda=8$ at $e^{-\beta}=2^{-10}$, and compared with nine other scalable methods from computer science~\cite{festa_algorithm_2001,simpson_efficient_2016}. Both belief propagation algorithms obtained the smallest number of violations; ours is somewhat faster, perhaps because we use a continuous model and low-degree polynomials. 

We turn next to parameter estimation. If we do not know how noisy comparisons are, we should use the data to estimate $\beta$.
The total probability that a model will generate $G$, averaged over all permutations $\bm{\pi}$, is 
\begin{equation}
\begin{aligned}
	P(G) &=  {1 \over n!} \sum_{\bm{\pi}} \prod_{(i,j)\in G} f(\pi_i, \pi_j) \\
	&= \int_{[0,1]^n} \prod_{(i,j) \in G} \frac{e^{-\beta \Theta(x_i-x_j)}}{1+e^{-\beta}} = Z(\beta)
\end{aligned}
\end{equation}
Using the thermodynamic relation $Z = e^{S - U}$ we approximate $Z(\beta)$ from the Bethe entropy and the energy, where the cavity approximation for the energy is
\begin{equation}
\label{eq:ubethe}
	\UBethe = -\sum_{(i,j) \in G} \int_{0}^1 \int_0^1 \mu_{ij}(x,y) \ln f(x,y) \,\dx\, \mathrm{d}y  . 
\end{equation}
If we have no prior information about $\beta$, we can determine its most-likely value by maximizing $Z(\beta)$. 

Finally, we discuss model selection. How can we choose between two different model classes, i.e., two different Hamiltonians? We might want to choose between the step-function model, where the probability that $i \prec j$ or $i \succ j$ depends only on the relative order of these two items, and a model where these probabilities depend on how far apart they are, for example the Bradley-Terry-Luce (BTL) model~\cite{bradley_rank_1952} (which actually dates back to Zermelo~\cite{zermelo_berechnung_1929}), or SpringRank~\cite{de_bacco_physical_2018}.
Let us consider BTL---a popular model of user preferences, similar to Elo Chess ratings where the probability a user prefers $i$ to $j$ or that $i$ will beat $j$ in a chess game is a logistic function
\begin{equation}
f(x_i,x_j) = \frac{e^{\beta(x_i-x_j)}}{e^{\beta(x_i-x_j)} + e^{\beta(x_j-x_i)}} = \frac{r_i}{r_i+r_j}
\label{eq:btl}
\end{equation}
where $r_i = e^{2\beta x_i}$. In general the continuous ranks in this model are allowed to range over the real line. We can scale them to the unit interval, and thus use our Chebyshev approximation, by varying $\beta$. This corresponds to assuming a uniform prior of width $\beta$ on the $x_i$.
To analyze the model we simply replace 
$\Theta(x-y)$ with $x-y$ in the belief propagation equations, Eq.~\eqref{eq:BP_message_general}.
We can use the orthogonality of the Chebyshev polynomials to efficiently compute the integral as a matrix product.

Given the observed comparisons, which of these two models should we prefer?
Both the models have one free parameter, and so a simple approach is to prefer the model with the largest maximum likelihood.

Using our methods, we computed the maximum likelihood values for ATP tennis tournament matches for the ten years $2010$--$2019$~\cite{jeff_sackmann_atp_nodate}. On average there were $n=435$ players. The average degree, i.e., the average number of games each player played it, was number of interactions $\lambda=11.0$. We found, interestingly, that the BTL model is only preferred over the step function model in one of the ten years ($2010$). In other years, the data is better explained by the step function model---where the probability that a weaker player beats a stronger one is fixed, rather than depending on the difference in their ranks.

To summarize, we have shown how natural problems involving permutations, rankings, and orderings can be treated as continuous spin systems. This includes counting linear extensions of a partial order, inferring the order in which nodes joined a growing network, and finding minimum-violation rankings. We discussed both models where probabilities depend only on the ordering, and those such as the Bradley-Terry-Luce model where they depend on differences in rank. We derived an efficient belief propagation algorithm using low-degree polynomials to compute marginals and entropies, and found that it is accurate on both sparse random graphs and some graphs with short loops. By using the Bethe free energy as an estimate of the log-likelihood, it can also perform parameter estimation and model comparison, and can be readily applied to real-world data.

\begin{acknowledgments}
\emph{Acknowledgments.} This work was supported by NSF grant BIGDATA-1838251. We thank Jiaming Xu and Jean-Gabriel Young for helpful conversations. Code implementing our methods is available at \url{https://github.com/gcant/pairwise-comparison-BP}.
\end{acknowledgments}

\bibliographystyle{numeric}
\bibliography{permutations}

\end{document}